\documentclass[10pt, a4paper,twocolumn]{article}
\usepackage[utf8]{inputenc}
\usepackage[T1]{fontenc}
\usepackage{fixltx2e}
\usepackage{graphicx}
\usepackage{longtable}
\usepackage{float}
\usepackage{wrapfig}
\usepackage{rotating}
\usepackage[normalem]{ulem}
\usepackage{amsmath}
\usepackage{marvosym}
\usepackage{wasysym}
\usepackage{amssymb}
\usepackage{lrec}
\usepackage{url}
\usepackage{covington}
\author{Serge Sharoff}
\date{}
\title{Know thy corpus! Robust methods for digital curation of Web corpora}

\name{Serge Sharoff}
\address{
  Centre for Translation Studies, \\
  University of Leeds, LS2 9JT, UK\\
{\small \texttt{s.sharoff@leeds.ac.uk}}
}

\abstract{This paper proposes a novel framework for digital curation of Web corpora in order to provide robust estimation of their parameters, such as their composition and the lexicon.  In recent years language models pre-trained on large corpora emerged as clear winners in numerous NLP tasks, but no proper analysis of the corpora which led to their success has been conducted.  The paper presents a procedure for robust frequency estimation, which helps in establishing the core lexicon for a given corpus, as well as a procedure for estimating the corpus composition via unsupervised topic models and via supervised  genre classification of Web pages.  The results of the digital curation study applied to several Web-derived corpora demonstrate their considerable differences.  First, this concerns different frequency bursts which impact the core lexicon obtained from each corpus.  Second, this concerns the kinds of texts they contain.  For example, OpenWebText contains considerably more topical news and political argumentation in comparison to ukWac or Wikipedia.  The tools and the results of analysis have been released.
\\ \newline
}
\begin{document}

\maketitleabstract

\section{Introduction}
\label{sec-1}
The vast amount of text data available from the Web provides a very good window for looking into a lot of language at once \cite{sinclair91}, which is extremely useful for developing language resources.  For example, unlabelled data from the Web are often used to pre-train embeddings for further downstream applications.  Many pre-trained language models, such as ELMO, BERT or GPT, have been produced by taking Web resources, such as the 1B Word Benchmark for ELMO, Wikipedia with the Toronto Book Corpus for BERT or WebText for GPT.  However, the impact of using a particular Web-derived corpus on the outcomes of pre-training is not clear, since corpora obtained from the Web lack curated categories, while it is known that such factors as the domain and quality of the training set are crucial to the performance of the model.  Digital curation of Web resources makes them similar in spirit to what has been used in traditional corpora, such as the BNC, as this allows estimation of its implications as well as drawing comparison to other Web corpora.  

One such issue is obtaining reliable estimates of the probability of occurrences for words, n-grams and linguistic constructions, since raw frequency counts are prone to bursts.  Adam Kilgarriff referred to this as a "whelk" problem \cite{kilgarriff97-frqc}: if a text is about whelks, no matter how infrequent this word is in the rest of the corpus, it's likely to be in nearly every sentence in this text.  If a corpus has a small bias towards texts of a particular kind, words related to its topics can get unreasonably high raw frequency at the expense of other words which can be more reasonably considered as belonging to the core lexicon.  

The following two extracts of word sequences are from the ranked frequency list of the OpenWebText corpus:\footnote{Here and below the subscripts indicate the rank of the respective word in frequency lists.}

\emph{down$_{\text{138}}$, \textbf{game}, say, same, against, why, \textbf{trump}, news, since, day, off, own, \textbf{government}, between} \ldots{}

\emph{determined$_{\text{2175}}$, favor, \textbf{license}, prepared, wonderful, combined, \textbf{stdclass}, percentage, tree, entry, \textbf{feed}, vast}

The frequencies of highly topical words, like \emph{game, Trump, government}, are close to those of function words; similarly \emph{stdclass} is unlikely to belong to the core lexicon.  This suggests that there are unknown biases in OpenWebText.  Many Deep Learning approaches need to limit their lexicon by applying frequency thresholds, since neural predictions need to choose from a relatively small number of options, such as 20-30,000 words.  Even methods operating with subwords, such as BPE \cite{sennrich16subw}, produce a substantial proportion of full-word entries.  For example, out of 22,702 BPE codes of the uncased BERT model 20,079 items are full words, which are taken from the frequency list of the respective training corpus.  

Another important problem with corpora derived from the Web concerns the lack of reliable information about their composition.  Usually Web corpora consist of millions of Web pages retrieved as a result of making queries to the search engines \cite{sharoff06ijcl}, crawling from a seed list \cite{baroni09} or taking URLs which the users shared through social media, such as reddit or Twitter \cite{radford19}.  Their composition can vary because of the procedure for their construction and because of the pipeline for their processing to obtain the textual contents and to remove the duplicates \cite{pomikalek12}, so that the results obtained from pre-training on their basis need to be validated.

The key contribution of this paper consists in proposing methods for analysing a large Web corpus with the aim of understanding its composition and its biases and for determining its core lexicon by robust estimation of the expected probabilities.  The structure of the paper is as follows:
\begin{itemize}
\item robust methods for detecting frequency bursts;
\item estimation of the topical composition via topic models;
\item estimation of the genre composition via supervised classification.
\end{itemize}

\section{Corpora}
\label{sec-2}
\begin{table}[htb]
\centering
\begin{tabular}{lllll}
 & Texts & Words & Lexicon & L10\\
\hline
OpenWebText & 8653K & 8706M & 31189K & 2425K\\
ukWac & 2542K & 1875M & 6286K & 832K\\
Wikipedia & 2524K & 1242M & 8168K & 1010K\\
\end{tabular}
\caption{Corpora used for experiments \label{tabCorpora}}
\end{table}

This study presents methods for analysing corpus anatomy on the basis of three Web corpora commonly used for pre-training.  OpenWebText is a public replication of the corpus used to train GPT-2 \cite{radford19}, which is based on extraction of Web pages from all URLs upvoted 3 or more times on the Reddit website.  OpenWebText (OWT) is also one of the components used for pre-training Roberta \cite{liu19roberta} using a publicly available pipeline.\footnote{\url{https://github.com/jcpeterson/openwebtext}}  In contrast to extraction of popular URLs, ukWac is a corpus produced by crawling the \verb~.uk~ Internet domain \cite{baroni09}.   Wikipedia is also often used for pre-training, in particular in such commonly used models as BERT \cite{devlin18} and fastText \cite{joulin17}.   The size of the lexicon for each corpus in Table \ref{tabCorpora} is presented for all orthographic tokens (excluding punctuation and tokens only consisting of numbers), as well as the lexicon of tokens occurring 10 or more times (L10).

Another important difference between these corpora is that there are many users authoring any given text in Wikipedia, while a single author is typically responsible for writing each text in other Web corpora.

\section{Estimation of frequency distributions \label{secRobust}}
\label{sec-3}

\subsection{Notation}
\label{sec-3-1}
This study uses the following notation to describe the frequency distributions:

\setlength{\tabcolsep}{2pt}
\begin{center}
\begin{tabular}{lp{6cm}}
$c_i$ & the number of occurrences of a word in a text $i$\\
$n_i$ & the size of a text $i$ in tokens\\
$C=\sum c_i$ & count, the total number of occurrences of a word in a corpus\\
$R=\sum r_i$ & robust count over robust by-text frequencies, which are immune to frequency bursts (see below)\\
$N=\sum n_i$ & the number of tokens in a corpus\\
$T=\vert n_i\vert$ & the number of texts in a corpus\\
$p_i=\frac{c_i}{n_i}$ & the probability of a word in a text $i$\\
$\mu=\frac{\sum p_i}{T}$ & macro-average of by-text probabilities\\
\end{tabular}
\end{center}

The frequencies are counted within the boundaries of individual texts, since texts are normally written by a specific author in a specific genre on a specific topic, so they offer natural units of analysis for studying variations of word use. There are some statistical complexities introduced by focusing on whole texts in comparison to splitting corpora into equally sized parts, but this is the preferred form of analysis for this paper, because texts provide natural boundaries between topics.  

\subsection{Robust frequency estimation}
\label{sec-3-2}
Frequency of linguistic phenomena, e.g., how common a word or construction is overall or is expected to be in a new text, has been of interest to researchers even before the invention of the computers. In the beginning of the 1900s, Andrei Markov investigated the frequencies of n-grams in poetry \cite{hayes13}, the first proper frequency dictionary has been produced for German at the end of the 19$^{th}$ century \cite{kaeding1898}, which was followed in the middle of the 20th century by the General Service List for English \cite{west53} and frequency studies on the Brown Corpus \cite{kucera67}.  

In addition to language teaching and lexicographic applications, frequency lists are needed to produce the probability estimates in many NLP applications, such as Machine Translation, Information retrieval, Speech recognition, Text classification. A lot of attention in language modelling has been paid to estimating the frequency of \emph{unseen} n-grams, while the problem addressed in this section concerns reliable frequency estimates of \emph{known} words in the presence of frequency bursts.

This can be done by introducing elements of robust statistics, which restricts contributions from outlying observations. It is known that traditional frequency measures, such as the mean (an estimator of location) and standard deviation (an estimator of scale) are not robust to outliers: a single frequency burst can move them out of bounds. Therefore, the field of robust statistic has introduced several robust estimators \cite{rousseeuw93}.

A commonly used robust estimator of scale is Median Absolute Deviation ($MAD$): $$MAD=b \times median \left| x_i - median(x) \right|$$ i.e., taking the median of the absolute differences from the median of $x$. Rousseeuw \& Croux introduced another scale estimator $S_n$ with more attractive gross-error sensitivity properties in comparison to MAD:
$$S_n=c \times median_i \left( median_j (\,\left| x_i - x_j \right|\,) \right)$$

It is the median of pairwise differences in word frequencies across texts. The values of the normalising constants $b=1.48$ for $MAD$ and $c=1.19$ for $S_n$ are used to match the standard deviation value when the measures are applied to normally distributed data \cite{rousseeuw93}.

As for robust estimators of location, the most commonly used measure is the median. However, it completely ignores variation of values around the median item, for example, for skewed distributions it does not reflect the difference between the two tails. Research in robust statistics proposed other robust measures of location, such as Huber's M-estimator, which is based on the idea of taking the values of non-outlying items at their face value and discounting the effect of the items outside a pre-defined range \cite{wilcox12}. The procedure is iterative: it starts with $\mu_0=Median$ and updates $\mu_{k+1}$ by discounting the contribution of the items which satisfy the condition:

$$\left|x_i-\mu_k\right| > K \times MAD$$

One problem in direct application of robust measures to word frequency lists consists in the prevalence of zero frequencies, for example, merely 73 words occur in more than half of the documents of OWT. This leads to the zero values of $median$, $MAD$ and $huberM$ for word frequency distributions for the vast majority of words. The way of dealing with this issue is by using robust methods to detect outliers within \emph{non-zero} frequency documents and to use the traditional mean of the discounted frequencies (Winsorisation).

More specifically, $huberM$ and $S_n$ are computed for the distribution of probabilities over all documents in which a word occurs. After that, the natural frequencies in these documents are capped as:
$$r_i=min(c_i,n_i\times(huberM(p_i)+k\times S_n(p_i))$$

The commonly used values of the scale constants ($K=1.28$ and $k=2.24$) are derived from statistical considerations on the influence functions \cite{huber11}.  They can be tuned to reflect the nature of the frequency distributions in corpora and the desired effect, i.e. the smaller they are the higher is the penalty on frequency bursts.

Summing up the $r_i$ values gives an estimate of the robust frequency $R$ for a word in a corpus, which can be used for establishing the core lexicon in order to determine the BPEs less affected by the frequency bursts.

\subsection{Core lexicon estimation results}
\label{sec-3-3}

The effect of Winsorisation can be measured by using the log-likelihood (LL) score \cite{rayson00}, i.e., by comparing the 
original frequency counts against the robust frequency counts from the same corpus as follows:

$$LL=R \ln \frac{R}{E} + C \ln \frac{C}{E}; \text{where } E=\frac{C+R}{2}$$

Words affected by the frequency bursts in ukWac ordered by their LL score are:

\emph{insurance, shall, sudoku, search, fire, waste, library, hotel, tax, wedding, credit, language, loan, cancer, mortgage, surfing, replies, hms, mulder, nigritude}

The topical word \emph{nigritude} in ukWac is a remainder of a Search Engine Optimisation contest run in 2004, in which the aim was to win by having a contestant's page at the top of Google searches for a non-sensical phrase \emph{nigritude ultramarine}.  Many of these pages remained in 2005 when ukWac was crawled, while they do not contain meaningful text and they should not contribute to the frequency count for \emph{nigritude}.  Some of the demoted words are related to insufficient cleaning of  webpages from boilerplate text (e.g., \emph{search, language}), some to commercial promotion (\emph{insurance, hotel}), some to text extracted from tabular formats (\emph{HMS}, as repeated many times in a table for different vessels).

Unlike overall frequency correction measures such as Juilland's D \cite{gries08}, Winsorisation reduces frequencies only in selected documents.  For example, the word \emph{library} is distributed more or less uniformly across many ukWac texts.  However, when programming manuals refer to programming libraries repeatedly, this bursts its frequency.  In the end, robust estimation reduces its overall frequency from 375,084 to 277,385.  Similarly, the estimation of the frequency of \emph{cancer} is affected by its repetition in bibliography lists, in which it occurs in such contexts as \emph{Int J Cancer 1991;48:816-820}.

Words affected by the frequency bursts in OWT according to their LL score are:\\
\emph{trump, posted, china, google, tax, clinton, climate, oil, o, la, los, e, apple, iran, fixed, y, van, beer, e-mail, deprecated, drupal, stdclass}

This indicates their repetition in very specific topics, as well as the presence of a small number of webpages in Spanish which nevertheless promoted the frequencies of \emph{o, la, e, y}, etc to the top 1,000 words in the raw count of OWT.  In comparison to ukWac, the OWT raw list contains fewer frequency bursts for words obviously related to commercial promotion.

The BERT lexicon mostly consists of the 20,000 most frequent words from the Wikipedia corpus (in addition to subword units).  Its validation via robust frequency estimation discovers a range of word-level elements, which are affected by the frequency bursts:\\
\emph{pomeranian, montane, spurred, substrates, encompassed, italianate, prelate, attaining}

According to the robust frequency estimates, they are all outside of the 20,000 word limit for robust counts.  On the other hand, there are 422 words, which are missing in the BERT lexicon, as they are below the 20,000 threshold in the raw frequency list, but are above this threshold in the frequency list of robust counts, for example:

\emph{appraisal, arisen, augment, bureaucratic, culmination, cultivate, divergent, numeric, overt, prosecute}

These words cannot be also represented by the BPE codes in the BERT lexicon, for example, the only BPE codes available in BERT for \emph{culminate} are \emph{culminated} and \emph{culminating}.  Robust frequency estimation offers the possibility of improving the lexical coverage of pre-training models.

\section{Topic estimation}
\label{sec-4}

\subsection{Topic modelling}
\label{sec-4-1}

\setlength{\tabcolsep}{0pt}
\begin{table*}[htb]
\centering
\small
\begin{tabular}{ll}
OWT & \\
\hline
8.60\% & police, court, officers, county, sexual, incident, charges, crime, children, prison, investigation, accused, charged, victim, assault\\
8.13\% & women, men, kids, girl, parents, book, mother, yes, remember, self, black, wasn, child, friend, shit, isn, feeling, knew, father\\
6.14\% & housing, trade, tariffs, council, companies, federal, billion, workers, minister, economic, tax, industry, trump, cent, union, percent\\
5.21\% & trump, mueller, clinton, fbi, russia, committee, investigation, comey, administration, counsel, border, obama, senate, election\\
4.52\% & season, players, nba, games, teams, league, ball, coach, draft, points, roster, win, played, playing, injury, sports, rookie, basketball\\
3.78\% & trump, israel, palestinian, election, war, democratic, vote, migrants, politics, anti, jerusalem, minister, speech, conservative\\
3.31\% & electric, battery, car, model, design, light, engine, display, air, water, launch, cars, features, speed, kerala, rear, vehicle, feet\\
3.01\% & technology, platform, companies, industry, product, learning, design, upgrade, automation, users, ideas, skills, react, google\\
2.45\% & music, album, song, band, artists, pop, rock, tour, vinyl, singer, fans, sound, usd, musical, guitar, track, art, studio, record\\
2.26\% & games, xbox, players, steam, cards, deck, card, player, damage, switch, dragon, character, spoiler, reload, console, magic\\
\hline
ukWac & \\
\hline
10.01\% & music, band, album, songs, sound, love, rock, night, playing, live, guitar, jazz, radio, dance, tracks, sounds, password, pop, played\\
6.76\% & sector, investment, financial, companies, market, carers, industry, performance, strategy, standards, policy, corporate, organisation\\
5.15\% & conference, social, policy, approach, faculty, practice, knowledge, communication, understanding, groups, discussion, cultural\\
4.87\% & god, love, jesus, got, man, went, posted, father, feel, thing, thought, mother, tell, let, friends, mum, came, told, jewish, says, friend\\
4.54\% & students, learning, student, courses, teaching, skills, education, study, college, degree, academic, postgraduate, studies, modules\\
4.13\% & education, schools, funding, learning, charity, trust, social, voluntary, organisations, youth, partnership, skills, projects, fund\\
3.28\% & credit, loan, card, pay, loans, money, goods, vat, account, charges, property, terms, paid, costs, customer, charge, insurance\\
2.94\% & wales, cardiff, award, june, city, awards, john, event, director, conference, royal, bbc, north, west, theatre, tate, nokia\\
2.83\% & golf, facilities, village, fishing, enjoy, park, ski, restaurant, pool, town, holiday, tour, beautiful, sea, resort, beaches, restaurants\\
2.79\% & register, mail, fee, address, application, telephone, send, advice, fax, online, request, child, data, parents, enquiries, dfes\\
\hline
Wiki & \\
\hline
8.82\% & album, song, chart, band, track, vocals, guitar, charts, label, listing, billboard, studio, records, singles, release, video, singer\\
7.33\% & league, cup, goals, club, tournament, championship, round, football, teams, rugby, match, draw, finals, division, professional\\
4.74\% & book, story, novel, plot, man, you, mother, said, how, woman, we, father, tells, find, love, characters, even, young, finds, himself\\
4.35\% & species, genus, mm, described, description, distribution, brown, marine, endemic, dark, leaves, habitat, grey, genera, plant, length\\
4.33\% & village, population, census, locality, municipality, workers, villages, km, town, rural, literacy, township, demographics, females\\
4.28\% & episode, episodes, television, show, festival, ep, tv, awards, documentary, production, films, award, cast, producer, premiered, role\\
3.87\% & building, historic, church, listed, places, buildings, brick, roof, street, style, story, tower, designed, windows, stone, architecture\\
3.82\% & football, coach, basketball, conference, ncaa, mf, tournament, head, df, schedule, games, record, fw, league, division, nfl\\
3.27\% & election, party, votes, assembly, candidate, democratic, council, minister, parliament, politician, legislative, seats, vote, results\\
3.00\% & law, court, trump, police, president, rights, act, minister, security, justice, legal, foreign, political, affairs, case, committee\\
\end{tabular}
\caption{Largest topics for the Web corpora \label{tabTopics}}
\end{table*}

The primary parameter for assessing a corpus concerns its contents with respect to its topics.  Generation of topic models is based on Latent Dirichlet Allocation (LDA), which estimates the distribution of probabilities of keywords belonging to different topics as well as the proportions of documents over the same set of topics. This is an unsupervised procedure, in which the unknown distributions are derived in repeated approximations from the distribution of latent variables \cite{blei03}.  For each topic $\vec{\beta}_k$ ($k=1\ldots K$) the task is to obtain the distribution of the probabilities of keywords over topics:

\setlength{\tabcolsep}{2pt}
\begin{center}
\begin{tabular}{lcccccc}
 & nba & players & season & steam & \ldots{} & xbox\\
$\vec{\beta}_1$ & 0.007 & 0.009 & 0.015 & 0.000 & \ldots{} & 0.000\\
$\vec{\beta}_2$ & 0.000 & 0.010 & 0.002 & 0.010 & \ldots{} & 0.011\\
\end{tabular}
\end{center}

In parallel the model also estimates the distribution of the degree its documents $\vec{\theta}_d$ ($d=1\ldots T$) belong to topics:\\
$$\vec{\theta}_1: (0.783 \beta_1, 0.002 \beta_2, ... 0.122 \beta_K)$$
$$\vec{\theta}_2: (0.002 \beta_1, 0.550 \beta_2, ... 0.213 \beta_K)$$

Instead of hard clustering, topic modelling assigns each document to a vector of topics.  
An estimation for the relative proportion of topics in a corpus can be provided by summing up the vectors of topics over all documents.  The similarity between the topics across corpora can be assessed via the Jensen-Shannon divergence \cite{fothergill16} as follows:
$$D_{JS}(\vec{\beta_1}, \vec{\beta_2})=\frac{D_{KL}(\vec{\beta_1}, \vec{B})+D_{KL}(\vec{\beta_2}, \vec{B})}{2}$$

where D$_{\text{KL}}(\vec{x},\vec{y})=\sum x_i \ln(x_i/y_i)$ is the Kullback-Leibler divergence and $\vec{B}=\frac{\vec{\beta_1}+\vec{\beta_2}}{2}$.

The implementation used in this study is based on the Multicore LDA model \cite{rehurek10} with the number of topics in the final experiment fixed as $K=100$.

\subsection{Topic modelling results}
\label{sec-4-2}

Table \ref{tabTopics} lists the largest topics from the three corpora, as described by their keywords.  Given that the soft assignment of topics in a topic vector is a distribution of probabilities which can be interpreted as the degree a document belongs to each topic, the first column of this table shows the proportion of the sum of all document probability assignments per topic as an estimate of its presence in a corpus.

There are three large topics (music, finances, sport), which are present in all corpora; even two prominent sport topics are detected in Wikipedia, one for more popular American sports (American football and basketball) and another one for soccer, rugby, etc.  While politics is present as a prominent topic in all corpora, OWT is markedly skewed to the current news in the US by having three kinds of prominent topical news (Topics 1, 4 and 6 in the rank of their prominence in Table \ref{tabTopics}).  

The topics related to education (Topics 5 and 6) and government services (Topic 10) and are more prominent in ukWac, partly because of the relative ease of crawling the educational and government websites, which less often rely on Javascript and have fewer explicit anti-robot restrictions.  Another considerably large topic in ukWac concerns commercial promotion (Topic 7), which is related to the process of its construction via wide crawling, which is more likely to include commercial pages. 

Topics related to descriptions of fiction/film (Topic 3), biological species (Topic 4) and locations (Topics 5 and 7) cover a more substantial portion of texts in Wikipedia in comparison to other corpora, which is related to its function of knowledge distribution. 

In spite of the large size of corpora, the LDA estimation is reasonably efficient: on a 24-core computer cluster node it takes less than 10hr to select the keywords in 9 billion words of OWT and 18hr to detect its topics.

\setlength{\tabcolsep}{2pt}
\begin{table*}[htb]
\centering
\begin{tabular}{rl||rl||rl}
 & OWT &  & ukWac &  & Wikipedia\\
\hline
42.01\% & argument & 19.30\% & argument & 91.33\% & information\\
14.91\% & personal & 11.78\% & personal & 4.64\% & review\\
4.96\% & news & 11.05\% & promotion & 1.13\% & news\\
3.36\% & instruction & 7.87\% & instruction & 0.88\% & argument\\
3.08\% & argument/personal & 7.85\% & news & 0.82\% & academic\\
2.75\% & review & 7.76\% & information & 0.61\% & info/review\\
1.96\% & promotion & 5.18\% & review & 0.42\% & info/news\\
1.85\% & academic & 3.88\% & academic & 0.30\% & instruction\\
0.87\% & information & 1.70\% & fiction & 0.12\% & info/instruction\\
0.85\% & legal & 0.87\% & legal & 0.12\% & info/academic\\
\end{tabular}
\caption{Distribution of genres \label{tabGenres}}
\end{table*}

\section{Genre estimation}
\label{sec-5}
\subsection{Genre classification}
\label{sec-5-1}
Genre is another important parameter for estimating variation in texts, as it is known that a mismatch in genres between the training set of a model and its application to a text has a considerable impact on such tasks as Part-Of-Speech tagging or Machine Translation \cite{santini10genreintro}.  Unlike the topical variation, which can be observed from the keywords using an unsupervised approach, non-topical variation in genres is expressed via stylistic features, which are harder to interpret \cite{biber95}.  Thus this requires a supervised approach to determine the genre categories.

This study uses a compact annotation scheme, which has been shown as suitable for reliable genre annotation of an arbitrary Web text \cite{sharoff18genres}, see Table \ref{tabGenres} for the list of categories used.  It is known that automatic genre classification can be biased by keywords specific to the training corpus \cite{petrenz10}.  
Therefore, our genre classifier uses a mixed representation which is based on keeping the most common word forms and replacing less common words with their POS tags, see \cite{baroni06translationese}.    For example, a hybrid text expressing the functions of review and promotion:

\begin{example}
\textit{It won the SCBWI Golden Kite Award for best nonfiction book of 1999 and has sold about 50,000 copies.}
\end{example}

\noindent
converts into a mixed representation as

\begin{example}
\textit{It won the} PROPN ADJ NOUN NOUN \textit{for best} NOUN NOUN \textit{of} [\#] \textit{and has sold about} [\#] NOUN.
\end{example}

The specific Machine Learning model in this study is based on a bi-directional LSTM classifier \cite{yogatama17} with the attention mechanism \cite{liu16}.

\subsection{Genre classification results}
\label{sec-5-2}

Table \ref{tabGenres} presents the distribution of genres detected in the three corpora.
As Wikipedia is the prototypical example of reference materials for information purposes, its texts which are not classified as \textbf{information} are either false negatives or Wikipedia articles exhibiting some features of typical reviews\footnote{\texttt{ https://en.wikipedia.org/wiki/1776 (musical)}} or news reports.\footnote{\url{https://en.wikipedia.org/wiki/Abu Nidal Organization}}  

OWT is clearly skewed towards argumentative texts, also including hybrid texts such as a combination of personal blogs with argumentation, while ukWac contains considerably more promotional texts.   The difference between the typical genres of OWT and ukWac can be explained by the methods of their collection: links to Web pages upvoted three or more times in reddit for OWT and crawling of the \verb~.uk~ domain for ukWac.  In the end OWT contains more links to discussion forums, opinion columns, political blogs and other argumentative texts.  At the same time, ukWac contains more advertising and shopping pages coming from a random snapshot of crawling.  Since OWT consists of pages upvoted by several users it is less likely to contain pages aimed at promotion.

\section{Related studies}
\label{sec-6}
Kenneth Church investigated the impact of frequency bursts on the probability of words by splitting a text into two parts ('history' and 'test').  He demonstrates a much greater probability of seeing a word in the test part once it occurred in the history part \cite{church00}. In the end, if the probability of seeing a topical word in a text once is $p(k=1)$, then the probability of seeing it twice is $p(k=2)\approx p/2$ rather than $p^2$ as expected in the binomial distribution.  Cf. also a more in-depth discussion by Harald Baayen in Chapter 5 of \cite{baayen01}. 

In his Spanish frequency dictionary Juilland introduced a measure of dispersion of word frequencies, which is essentially based on the standard error of the mean normalised by the mean \cite{juilland64}: 
$$D=1-\frac{\sigma}{\mu\sqrt{T-1}} \label{eqD}$$

This proposal was followed by several other measures aimed at identification and mitigation of such bursts, e.g., Carroll's, Rosengren's, Engvall's measures, see an overview in \cite{gries08}. Because of the inadequacies of these measures, Gries has also suggested his own measure, Deviation of Proportions (DP), which is defined as:\footnote{This can be followed by normalisation to ensure that its value is within $[0,1]$} 
$$DP=\frac{\sum |\frac{c_i}{C}-\frac{n_i}{N}|}{2} \label{eqDP}$$

More burstiness measures have been suggested by Katz with the aim of
using them in speech recognition, information retrieval and terminology
detection \cite{katz96}:

\setlength{\tabcolsep}{2pt}
\begin{center}
\begin{tabular}{lp{5cm}}
$p(k=0)=p_0$ & probability of no occurrences of the term in a text\\
$p(k=1)=p_1$ & probability of a single occurrence of the term in a text\\
$p(k\ge 2)=\sum p_r$ & probability of multiple occurrences of the term in a text ($r\ge 2$)\\
$\alpha = 1-p_0$ & proportion of texts containing the term\\
$\gamma = 1-\frac{p_1}{1-p_0}$ & proportion of 'topical' texts for the term\\
$B=\frac{\sum rp_r}{\sum p_r}$ & topical burstiness parameter\\
\end{tabular}
\end{center}

$\alpha$ shows how likely the word is to occur in a text irrespectively of the number of times it occurs there; $\gamma$ shows how likely it is to be used 'topically' (i.e. more than once within a text); and $B$ shows how intensely, on average, the word is used when it is used topically.

$\alpha$ is effectively the proportion of texts in which a word occurs, also it is the same as Engvall's measure \cite{gries08}. This measure is also directly linked to the IDF (Inverse Document Frequency) measure. By this count \emph{wonderful} becomes more common than \emph{stdclass}.  However, the application of range-based frequency lists is also limited by the fact that they do not distinguish evidence coming from short and long texts, so that their values can vary radically between otherwise reasonably similar corpora.

Language modelling pays attention to smoothing, i.e., estimating the frequency of 'unseen' n-grams, while the frequency of observed n-grams is measured as it is without using information from the document frequencies, only the sentence frequencies are sometimes taken into account \cite{heafield13}. Therefore, LM does not distinguish between the probabilities of \emph{new stdclass} vs \emph{wonderful moment}, which have similar raw frequencies (in OWT) and very different burstiness properties.

Another problem which concerns all of these measures is that we do not have an estimation of what reliable counts are likely to be: we can detect the lack of a well-behaving distribution across a number of documents, but this does not help in detecting the \emph{expected} frequency value. A common practice in frequency dictionaries is to multiply the raw counts by a dispersion measure (by Juilland's D in the frequency dictionaries), but this applies a uniform correction measure to the overall count, while the frequency bursts are specific to individual texts.

Active research in comparing corpus composition using keywords and most frequent words has started since \cite{kilgarriff01-comparing},  followed by \cite{kilgarriff12}.  It has been shown that the use of topic modelling helps in finding the differences between the Web corpora \cite{fothergill16,sharoff13bucc}.  Since the arrival of machine learning methods in the 1990s, genre classification and related approaches to classification of texts with respect to their stylistic features developed from \cite{karlgren94} to \cite{pritsos18}, see a recent overview in \cite{argamon19}.  However, genre classification methods have not been yet applied to very large corpora from the Web.  

\section{Conclusions and further work}
\label{sec-7}
The study explored the significant differences in the lexicon and in the composition of OpenWebText, ukWac and Wikipedia, three large Web corpora, which are commonly used for pre-training language models.  The size of a corpus (all of them measure in billions of words) is not the only consideration for its effective use in pre-training.  Its lexicon and its composition in terms of topics and stylistic properties are likely to impact the use of pre-trained models in downstream tasks.  The robust lexicon estimation and topic modelling are based on unsupervised methods, while the genre estimation task is based on supervised methods (using bi-LSTM and attention).  There is also a difference in the underlying representation, lexicon vs stylistic features using POS tags for the genre classifier.

The analysis framework based on these parameters provides three perspectives to describe a corpus.  In the study of the three corpora reported in the paper they all converged to consistent descriptions of each corpus.  First, the OpenWebText corpus is very topical, it is related to the current political situation, such as three prominent topics related to \emph{Trump}, because of the nature of its collection from upvoted links on social media.  From the viewpoint of its genre composition, OpenWebText was found to contain a much larger proportion of argumentative texts, such as opinion columns or argumentative blogs.  Also, OpenWebText emphasises the perception aspect of language use by collecting widely shared texts, while ukWac and Wikipedia contain a snapshot of available texts without taking into account the size of their audience.  This introduces other biases in their lexicon and composition, such as prominent topics related to rare plants, animals and locations in Wikipedia or texts from educational and government websites in ukWac.  From the viewpoint of its genre composition ukWac was also found to contain a larger proportion of texts aimed at commercial promotion, which is also shown through the frequency bursts and topic models.  

Apart from information about these specific corpora, this study contributes a framework for discovering such biases to make a suitable decision in using a particular corpus for pre-training.
The procedure presented in this study provides the basis for assessing how close a large corpus is to a specific application domain for pre-training.  Judging from the results of corpus composition, both Wikipedia (used for BERT) and OWT (used for GPT and Roberta) have very specific biases, which are beneficial to certain tasks, such as wide general knowledge for Wikipedia and processing of argumentative texts for OWT.  At the same, a corpus produced by Web crawling (such as ukWac, but also Open Crawl) is less biased, but more affected by spam.
The composition assessment procedure also enables selection of suitable subsets covering specific topics and genres, so that pre-training can be performed on a smaller corpus more suitable for a specific task.  
The classification framework used in this study is available under a permissive license.\footnote{\url{https://github.com/ssharoff/genre-keras}}

    In addition to detecting the corpus composition, this study proposes a method for building the core lexicon for a corpus, which overcomes the worst frequency bursts.  Robust frequency estimation offers the possibility of improving the lexical coverage of pre-trained models.  One key take-home message of this study is that frequency estimation for known words should not rely on raw counts.  It is not safe to estimate the probability of seeing a word as $p=\frac{C}{N}$.  Otherwise, it is easy to infer that $p(\text{\textit{stdclass}})=p(\text{\textit{wonderful}})$.  A better approach is to obtain a more robust frequency list by reducing frequency bursts.  The proposed mechanism is based on Winsorising the raw frequencies within documents using $huberM$ and $S_n$ values, it is effective in detecting frequency bursts.  For example, this mechanism demotes 422 frequency bursts from the main lexicon of BERT, replacing words like \emph{pomeranian} or \emph{montane} with promoted words such as \emph{appraisal} or \emph{divergent}.  
The frequency estimation tools are also computationally efficient in comparison to bootstrap or Monte Carlo methods.  When the document-level frequencies for words are available, the estimation takes less than 3 hours for any corpus reported in Table \ref{tabCorpora}.  The tools for frequency estimation are available under a permissive license.\footnote{\url{https://github.com/ssharoff/robust}}

\section{Acknowledgements}
\label{sec-8}
This work was undertaken on ARC3, part of the High Performance Computing facilities at the University of Leeds, UK.  The rationale behind this investigation was discussed with Adam Kilgarriff.  Also thanks to the anonymous reviewers.  As usual, I'm responsible for any remaining errors.

\section*{Bibliographical References}

\bibliography{bibexport}
\bibliographystyle{lrec}
\end{document}